\documentclass{article}
\usepackage[utf8]{inputenc}
\usepackage{graphicx}
\usepackage{amsmath}
\usepackage{amssymb}
\usepackage{amsfonts}
\usepackage{float}
\usepackage{array}
\usepackage{amsthm}
\usepackage[margin=2.5cm]{geometry}
\usepackage{authblk}
\usepackage[hidelinks]{hyperref}
\hypersetup{
    colorlinks=true,
    linkcolor=blue,
    urlcolor=blue,
    }
\usepackage[title]{appendix}
\usepackage{pdfpages}
\usepackage{listings}
\usepackage{float}

\numberwithin{equation}{section}

\usepackage[
    backend=biber,
    natbib=true,
    style=numeric-comp,
    sorting=none,
    maxnames=2,
    giveninits=true,
    url=true,
    arxiv=pdf,
    doi=true,
    eprint=true,
    isbn=true,
    date=terse,
    block=ragged,
    maxbibnames=99
]{biblatex}
\title{
  \includegraphics[width=0.4\textwidth]{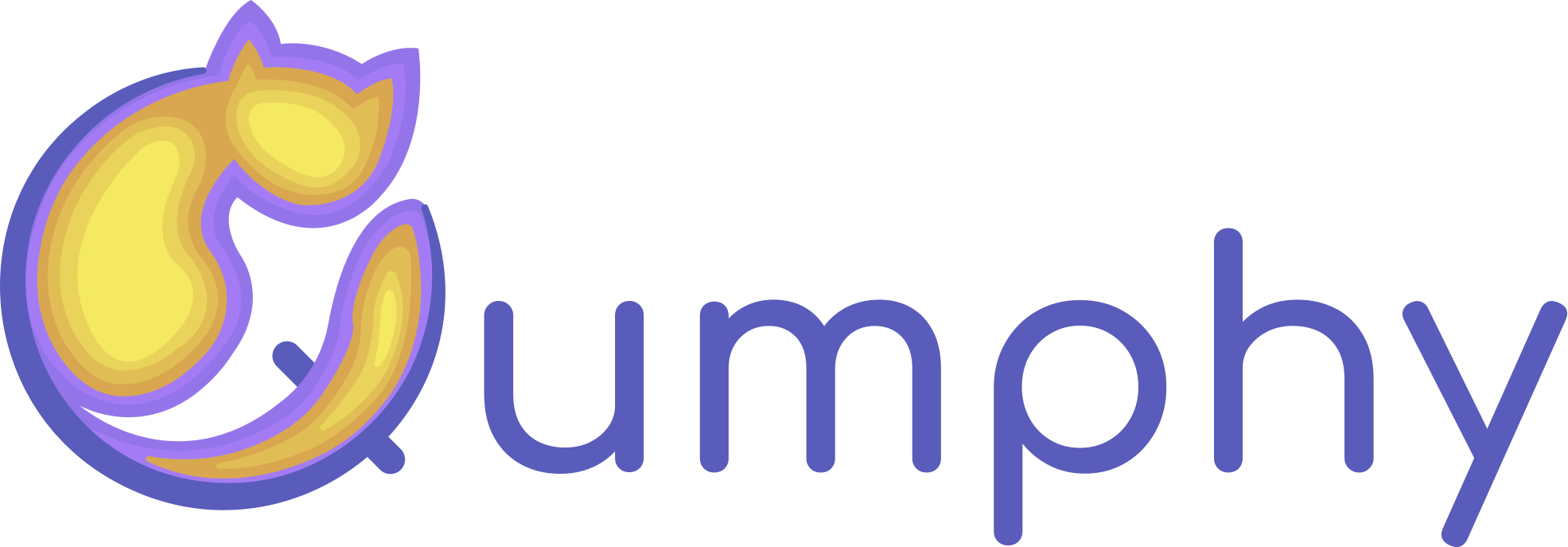}
  \hfill
  \includegraphics[width=0.4\textwidth]{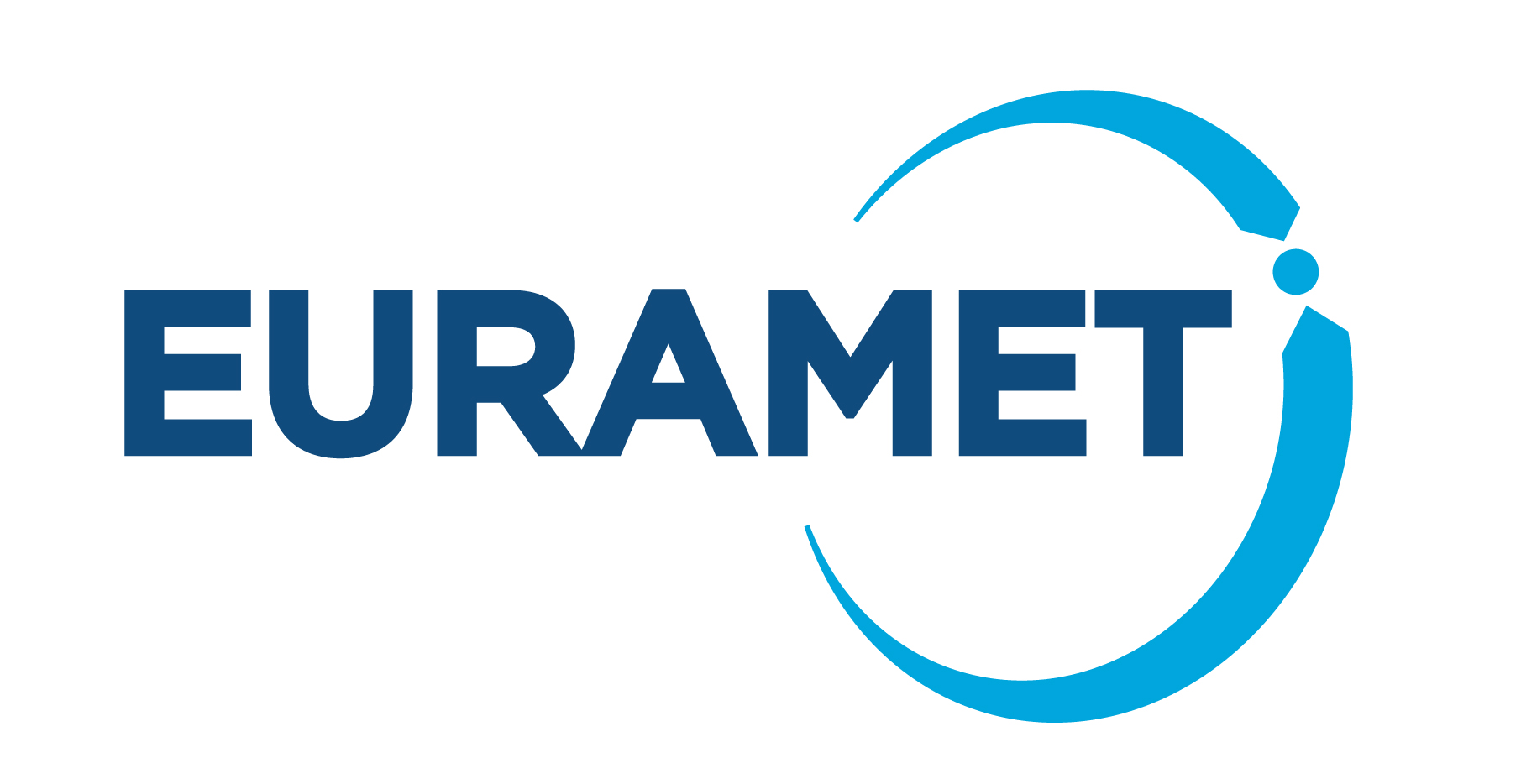} \\ [1.5cm]
  Benchmark Problems and Benchmark Datasets for the evaluation of Machine and Deep Learning methods on Photoplethysmography signals: the D4 report from the QUMPHY project}
\author[1, 2]{Urs Hackstein}
\author[3]{Jordi Alastruey}
\author[4]{Philip Aston}
\author[4]{Ciaran Bench}
\author[5]{Peter H. Charlton}
\author[6]{Loic Coquelin}
\author[7]{Nando Hegemann}
\author[8]{Vaidotas Marozas}
\author[9]{Mohammad Moulaeifard}
\author[3]{Manasi Nandi}
\author[8]{Andrius Petrenas}
\author[7]{Oskar Pfeffer}
\author[8]{Mantas Rinkevicius}
\author[8]{Andrius Solosenko}
\author[9]{Nils Strodthoff}
\author[3]{Sara Vardanega}
\affil[1]{corresponding author: urs.hackstein@lse.thm.de}
\affil[2]{Technische Hochschule Mittelhessen - University of Applied Sciences,  Giessen, Germany}
\affil[3]{King's College London, London, United Kingdom}
\affil[4]{National Physical Laboratory, Teddington, United Kingdom}
\affil[5]{University of Cambridge,
Cambridge, United Kingdom}
\affil[6]{Laboratoire National de Métrologie et d’Essais, Paris, France}
\affil[7]{Physikalisch-Technische Bundesanstalt, Berlin, Germany}
\affil[8]{Kaunas University of Technology, Kaunas, Lithuania}
\affil[9]{Carl von Ossietzky Universität Oldenburg, Oldenburg, Germany}
\date{\today}
\begin{document}

\maketitle  %
\section*{Summary}%
\label{sec:summary}
This report is part of the Qumphy project (22HLT01 Qumphy) that is funded by the European Union and is dedicated to the development of measures to quantify the uncertainties associated with Machine Learning algorithms applied to medical problems, in particular the analysis and processing of Photoplethysmography (PPG) signals. In this report, a list of six medical problems that are related to PPG signals and serve as Benchmark Problems is given. Suitable Benchmark datasets and their usage are described also.

\clearpage
\tableofcontents

\clearpage
\section{Introduction}%
\label{sec:introduction}
\subsection{Photoplethysmography}
Photoplethysmography (PPG) signals are rich in information and easy to measure passively without any physical or mental limitations of the subject. PPG is a widely used physiological sensing technigue. It consists of shining a light (often red, green or infrared) onto the skin and measuring the amount of light either reflected from, or transmitted through, a region of tissue.\\

\begin{figure}[h]
    \centering
\includegraphics[width=12cm]{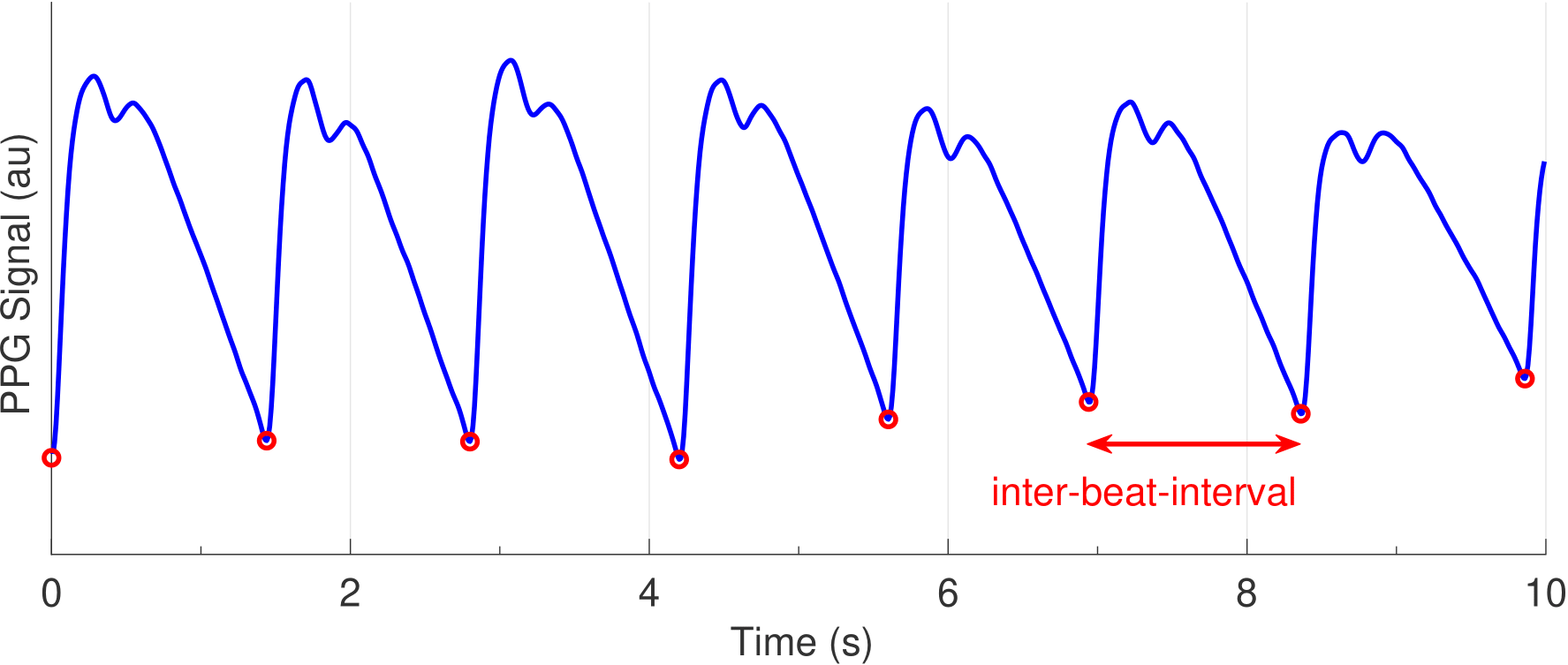} 
\caption{An exemplary photoplethysmogram (PPG) signal showing a pulse wave for each heartbeat. Pulse onsets, representing individual heartbeats, are shown as red circles. An inter-beat interval is labelled, corresponding to the time between consecutive heartbeats. Adapted from \cite{Charlton2020} by Peter H. Charlton - Own work, CC BY 4.0, \url{https://commons.wikimedia.org/wiki/File:Photoplethysmogram_(PPG)_signal.png}}
\label{ppgsignal}
\end{figure}

This amount varies with each heartbeat and the PPG signals measure the fluctuations in blood volume which occur with each heartbeat, see Figure \ref{ppgsignal}.

Photoplethysmography (PPG) signals contain valuable information on the cardiovascular, respiratory, and autonomic nervous systems which is not yet routinely exploited. They are popular as they are easy to obtain non-invasively and PPG devices are cheap and widely available.\\ 

Until today however, an algorithmic evaluation of PPG signals to infer physiological parameters or detect diseases is crucial for saving patients’ lives, but almost never used in clinical environments. One of the major reasons for this is the lack of trust in the output of any such algorithms.\\

Due to the vast amount of collected data, machine learning methodologies are essential for the extraction and evaluation of key features used for diagnosis. When applying machine learning in a medical context, however, confidence in the performance and predictions of the algorithms is particularly crucial since diagnostic mistakes can be fatal (false negative) or result in unnecessary anxiety and detrimental overtreatment (false positive). Hence an analysis of the uncertainty associated to ML algorithms and their predictions is indispensable to provide critical information about the quality and trustworthiness of the results produced.

\subsection{The Qumphy project}
The goal of the Qumphy project that is to satisfy these needs by developing an environment, i.e., a good practice guide including a software framework for independent assessment of accuracy and uncertainty of ML algorithms and benchmark cases to test and compare ML algorithms against, to increase trust in ML applications for PPG signals and lay a foundation towards standardisation of ML in healthcare. This project (22HLT01 QUMPHY) has received funding from the European Partnership on Metrology, co-financed from the European Union’s Horizon Europe Research and Innovation Programme and by the Participating States. Funding for the UK partners was provided by Innovate UK under the Horizon Europe Guarantee Extension.

\subsection{Aim of the report}

In this report we report that more than 5 datasets using real and/or synthetic PPG data have been generated that can be used to benchmark the accuracy and uncertainty of supervised machine learning and deep learning models. This includes making six reference problems and their respective datasets available to the medical device and digital health communities.

The following six reference problems are chosen as benchmark problems:
\begin{itemize}
    \item[1] Determine systolic and diastolic blood pressure
    \item[2] Detection of Atrial Fibrillation
    \item[3] Classification of Hypertension
    \item[4] Determine vascular age
    \item[5] Detection of Sleep Apnea
    \item[6] Determine respiratory rate
\end{itemize}

Our selection is based on the following aspects that we considered:

\begin{itemize}
    \item[(i)] How important is the problem to society?
\item[(ii)] Are there alternative methods not involving machine learning that are already established and sufficient?
\item[(iii)] How widely has the problem been tackled?
\item[(iv)] Are there sufficiently large, open datasets available?
\item[(v)] Is there a plausible physiological mechanism by which one could estimate the parameter from a PPG signal?
\item[(vi)] Is the data annotated?
\item[(vii)] How accurately can the ground truth be quantified?
\end{itemize}

In the following, we give a description of each individual benchmark problem together with their benchmark datasets and explain the possibilities to use them.

\section{Benchmark I: Determine systolic and diastolic blood pressure}%
\label{sec:benchmark_1}

\subsection{The problem}

\begin{figure}[ht] 
	\centering
	\includegraphics[width=12cm]{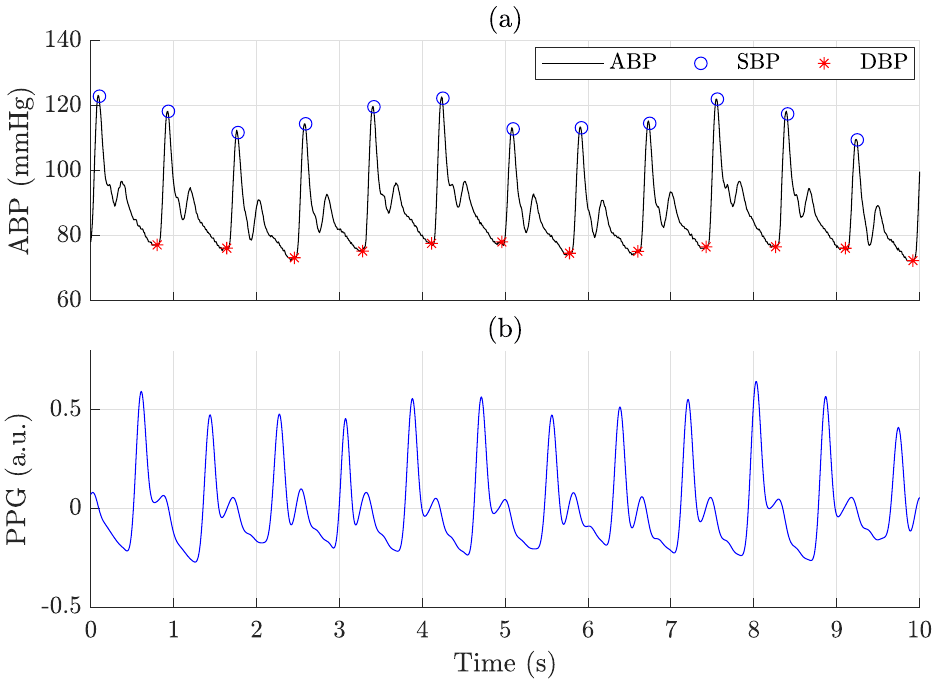} 
	\caption{The example of (a) arterial blood pressure (ABP) segment with labeled fiducial points (systolic (SBP) and diastolic (DBP) blood pressure) and (b) PPG segment.}
	\label{ABP_PPG_Vital}
\end{figure}

Blood pressure (BP) is one of the most commonly taken physiological measurements as it can be used to monitor cardiac health and predict adverse cardiac events, and is essential for the selection and monitoring of antihypertensive (BP lowering) treatments \cite{ESCguidelines}. Blood pressure consists of two measurements, namely  the peak systolic blood pressure (SBP) and the trough diastolic blood pressure (DBP) (see Figure \ref{ABP_PPG_Vital}). These are most commonly measured using an inflatable cuff, which has many limitations and which precludes continuous monitoring. Accurate blood pressure estimation from PPG signals is a regression machine learning problem and would have the advantage of generating a continuous readout of blood pressure, thus providing much extra, valuable information for the clinician.

\subsection{Potential datasets}

This task requires datasets which include PPG segments together with (segment averaged) systolic and diastolic blood pressure measurements. There are quite a few such datasets either publicly available, available on request, or private. The main disadvantage of many of these, from a machine learning perspective, is that they are relatively small. 

Another important factor to consider is the setting for collection of the data. The largest datasets are typically collected in clinical settings as it is relatively easy to collect data from patients in hospital. The disadvantage of such datasets is that these are sick patients who are undergoing medical interventions and are often on medication. One such dataset is MIMIC-III \cite{MIMIC} which has data collected from over 40,000 patients who were in intensive care. We note that an even larger dataset, MIMIC-IV is now available \cite{MIMIC}. The PPG signals in these datasets are generally collected from a pulse oximeter placed on a finger. Alternatively, data can also be collected in a community setting from generally healthy subjects, although of course some of these will have ongoing medical issues and be on medication. Such data is generally collected from a smartwatch on the wrist (or similar wearable device). One of the largest such datasets is Aurora BP \cite{auroraBP} which contains data from over 1,000 participants and was collected with the aim of improving cardiovascular monitoring.

The length of signals available is another factor. The UK Biobank \cite{UKBiobank} has data available from over 200,000 subjects but the PPG signal is just a single beat, which is not ideal for machine learning. It is also not a free dataset. On the other hand, the signals in the MIMIC-III dataset range in length from minutes to several days, from which many shorter segments can be extracted.

We will consider in detail two datasets, namely Aurora BP \cite{auroraBP}, which is a large dataset collected in a community setting, and VitalDB \cite{PulseDB}, which is also a large dataset collected from surgery patients.

\subsubsection{Aurora BP}\label{AURORABP_dataset_presentation_section}

The \textbf{AuroraBP} dataset \cite{auroraBP} consists of PPG (and other) signals together with simultaneously collected auscultatory or oscillometric measurements of systolic and diastolic blood pressure. Measurements were made in a lab over a 24 hour period and some subjects also had ambulatory measurements. There were 1,125 particpants in the study (ages 21-85, average age $45.1\pm 11.3$, 49.2\% female, multiple hypertensive categories). Fitzpatrick skin tone class is provided for 823 participants. However, the distribution of skin tones is highly imbalanced, with over 90\% of classes 1, 2 or 3 and less than 1\% of class 6 \cite{Aston2024}. The PPG sensor was placed on the anterior surface of the arm (underside), in contrast to smartwatches which are worn on the posterier surface of the arm, and signals have been resampled to a frequency of 500\,Hz.

The dataset contains 38,623 waveform records of varying durations. The average signal length was 19.73\,s and the range was 9.1\,s$-$87.1\,s.

Requests for the data have to be made to the Data Access Committee and should include a research project description and details of the investigators.

The AURORA BP dataset has already been used in the following state-of-the-art studies \cite{cisnal_robust_2024, liu_hgctnet_2024}.

\subsubsection{Vital DB}

The PulseDB dataset \cite{PulseDB} is a selection of high quality segments from two datasets, namely VitalDB (2,938 subjects) and MIMIC-III Waveform Database Matched Subset (2,423 subjects). The \textbf{PulseDB Vital} subset consists of 2,938 non-cardiac surgery patients (2,938 structured MAT files), containing raw and filtered synchronized physiological signals: (i) electrocardiogram (ECG), (ii) finger photoplethysmogram (PPG), and (iii) reference invasive blood pressure measurements. The PPG signals were collected from the fingertip, which poses a drawback since the majority of wearables utilize reflection PPG sensors positioned on the upper part of the wrist. However, the newest type of wearables—smart rings—can record PPG signals from the finger \cite{kim_first_human_2023} and
may still benefit from the results obtained from fingertip PPG recordings.

The PulseDB Vital subset includes patients during perioperative periods, which involve surgical operations (general, thoracic, urological, and gynecological surgery). The patients underwent routine or emergency operations at Seoul National University Hospital. The PulseDB Vital dataset provides information of subject age, height, weight, and the body-mass index (23.07 $\pm$ 3.53 $\textrm{kg/m}^{2}$). The average age of patients is 58.76 $\pm$ 15.02 years (54.73\% male).

The PPG signals and continuous BP waveforms are sampled at a rate of 125 Hz. The signals are segmented into 10\,s windows, providing a single average value of systolic and diastolic blood pressure for each segment. All 10\,s segments related to PPG or BP signals with outliers or anomalies were removed. This means all aleatoric uncertainty was removed, and only epistemic uncertainty was left. The average duration of patient signals is 1.38 $\pm$ 1.18 h (range 10.08\,s$-$9.16\,h). Also, the PulseDB Vital data provides annotated fiducial points of ECGs (R peaks), PPGs (peaks and onsets) and BP measurements (peaks and onsets), which can be utilized to obtain beat-to-beat SBP/DBP sequences. In addition, version 2 of the Pulse DB Vital dataset provides the absolute time of each sample in the 10\,s segment, and raw/filtered ECG and PPG signals with non-normalized absolute amplitudes. The PulseDB Vital dataset is released under the Creative Commons Attribution-NonCommercial-ShareAlike 4.0 International (CC BY-NC-SA 4.0).

\subsection{Making the datasets available}

\subsubsection{Aurora BP}\label{AURORABP_access_section}

The Aurora BP dataset is not openly available. An application for access to the data should be made to the Data Access Committee. The process is described \href{https://github.com/microsoft/aurorabp-sample-data?tab=readme-ov-file#data-access}{here}.

\subsubsection{VitalDB}

The PulseDB dataset, which includes the VitalDB dataset, can be downloaded from \href{https://github.com/pulselabteam/PulseDB?tab=readme-ov-file}{here}. After downloading, you can either proceed with the original instructions involving MATLAB and Python codes to generate the .npy files or implement a few adjustments to create lighter files tailored to the QUMPHY project's objectives. The adjustments are outlined in the D1 report of the QUMPHY project and \cite{D1} as well in the accompanying \href{https://gitlab.com/qumphy/d1-code/-/blob/main/Preprocessing/PulseDB_Deepbeat_preprocessing/PulseDB_Preprocessing_Guide.md?ref_type=heads}{repository}. 

\subsection{Dataset usage}

In the following, an introduction to how to use these datasets for the problem of blood pressure estimation from PPG signals is given.

\subsubsection{Aurora BP}

Matlab code is provided which splits the Aurora BP dataset into 10 folds. The training, validation, test and calibration sets can then be defined by the user from these 10 folds as appropriate, with a split of 7/1/1/1 folds being suggested respectively (or 8/1/1 if a calibration set is not required). Using the folds for cross validation is clearly also an option. All the records for each subject are contained in a single fold. The folds have been stratified to give similar distributions of the following classes:
\begin{itemize}
\item 
Gender: Male/female.
\item 
Blood pressure class: Three classes are defined in terms of systolic blood pressure (SBP) and diastolic blood pressure (DBP), and in accordance with the 2024 ESC Guidelines \cite{ESCguidelines} by
\begin{itemize}
\item 
Non-elevated: $\mathrm{SBP}< 120$ mmHg and $\mathrm{DBP}< 70$ mmHg
\item 
Elevated: $120\leq\mathrm{SBP}< 140$ mmHg or $70\leq\mathrm{DBP}< 90$ mmHg
\item 
Hypertensive: $\mathrm{SBP}\geq 140$ mmHg or $\mathrm{DBP}\geq 90$ mmHg
\end{itemize}
Note that the Aurora BP variables \emph{baseline\_sbp} and \emph{baseline\_dbp} are used for the SBP and DBP values \cite{auroragitreference}. These provide a single value for each subject, which then allowed classification into one of the 3 classes.
\item 
Cardiovascular disease: The Aurora BP metadata contains information about a range of self-reported cardiovascular diseases. Two classes are used, namely no cardiovascular disease and cardiovascular disease if at least one disease is reported.
\item 
Body mass index (BMI): Three classes of BMI are used:
\begin{itemize}
\item 
Healthy: $\mathrm{BMI}<25$ kg/m$^2$
\item
Overweight: $25\leq\mathrm{BMI}<30$ kg/m$^2$
\item
Obese: $\mathrm{BMI}\geq 30$ kg/m$^2$
\end{itemize}
\end{itemize}

\subsubsection{VitalDB}

 The PulseDB dataset contains some metadata. The file \texttt{metadata.csv} comprises 1,494,474 rows and 20 columns, representing the combined MIMIC+VitalDB version of PulseDB. Key columns are as follows:
\begin{enumerate}
\item 
``source": Indicates the dataset origin (1 for the VitalDB dataset, 0 for the MIMC dataset).
\item
``bmi", ``weight", ``height": These metrics are available only for the VitalDB dataset.
\item
``dbp\_avg" and ``sbp\_avg": These represent the average diastolic and systolic blood pressures respectively, which can be used as prediction targets. For convenience, we also provide [``dbp\_avg",``sbp\_avg"] in the column ``label" as prediction target.
\item
``set": Differentiates the data into five distinct categories (see Table 4 of \cite{PulseDB}):
\begin{itemize}
\item
set = 0 for train set
\item
set = 1 for calibration based testing set
\item
set = 2 for calibration-free testing set
\item
set = 3 for AAMI calibration set
\item
set = 4 for AAMI testing set
\end{itemize}
\item
``set\_calib", ``set\_calibfree," and ``set\_aami": Provides train/validation/calibration/test splits for the three scenarios (where X is ``calib", ``calibfree" or ``aami"):
\begin{itemize}
\item
set\_X=0 for training
\item
set\_X=1 for validation
\item
set\_X=2 for calibration
\item
set\_X=3 for testing
\end{itemize}
\end{enumerate}

\textbf{Usage example}

To train a model on the VitalDB subset in the calibration free scenario, the following steps are required:
\begin{enumerate}
\item 
Load the file \texttt{metadata.csv}:
\begin{lstlisting}
    
   import pandas as pd
    #Load the CSV file
   df = pd.read_csv('metadata.csv')
\end{lstlisting}

\item
Select the desired indices based on the entries in the column ``set\_calibfree" with source=1 (VitalDB):
\begin{itemize}
\item 
Train on entries with set\_calibfree=0
\begin{lstlisting}
  indices_train = df[(df['set_calibfree'] == 0) & (df['source'] == 1)].index
\end{lstlisting}
\item 
Validate on entries with set\_calibfree=1 (e.g.\ for hyperparameter tuning and/or model selection)
\begin{lstlisting}
  indices_val = df[(df['set_calibfree'] == 1) & (df['source'] == 1)].index
\end{lstlisting}
\item 
Calibrate the model uncertainty with set\_calibfree=2 (e.g.\ for calibration or conformal prediction)
\begin{lstlisting}
  indices_cal = df[(df['set_calibfree'] == 2) & (df['source'] == 1)].index
\end{lstlisting}
\item 
Test model performance with set\_calibfree=3
\begin{lstlisting}
  indices_test = df[(df['set_calibfree'] == 3) & (df['source'] == 1)].index
\end{lstlisting}
\end{itemize}

\item
The selected row numbers for the subsets identified in the previous step can be used to extract the corresponding signals from the signals.npy file. The columns sbp\_avg and dbp\_avg can be used as prediction targets:
\begin{lstlisting}
  import numpy as np
  # Load the signal file.npy
  signals = np.load('signals.npy')
  # Extract signals corresponding to indices_train
  signals_train = signals[indices_train]
\end{lstlisting}

Note that if uncertainty quantification is not of interest, the validation and calibration sets can be combined by using
\begin{lstlisting}
    set_calibfree=1 & set_calibfree=2 
\end{lstlisting} 
to select validation set samples.
\end{enumerate}

\section{Benchmark II: Detection of atrial fibrillation}%

\label{sec:benchmark_2}
\subsection{Problem}

Atrial fibrillation (AF) is a major health and economic concern, reaching epidemic levels. More than 33 million people worldwide are diagnosed with AF, and this number is projected to double by 2050~\cite{krijthe2013projections}. Paroxysmal AF can be asymptomatic and difficult to detect, making early diagnosis crucial to prevent severe outcomes like ischemic brain stroke~\cite{martin20242024}. AF detection remains challenging due to its paroxysmal and sometimes brief episodes~\cite{van20242024}. Opportunistic pulse palpation followed by a 12-lead ECG is a recommended screening method for individuals over 65 but is ineffective for asymptomatic cases, as it is performed only when symptoms are reported. Therefore, developing technologies for screening high-risk populations in a convenient and affordable way is essential~\cite{jones2020screening}.\\ 

Traditional monitoring methods, such as ambulatory ECG monitors and cardiac event recorders, can be uncomfortable due to skin irritation and disruption to daily life, especially when long-term monitoring over several weeks is required to detect asymptomatic paroxysmal AF. Recent advances in wearable technology have introduced more convenient and cost-effective alternatives, such as PPG acquisition using a built-in camera of a smartphone~\cite{lee2012atrial}, a web camera~\cite{couderc2015detection}, earlobe sensor~\cite{conroy2017detection}, or a smart wristband~\cite{solovsenko2019detection}.\\

Finger- and wrist-based PPG face challenges from artifacts~\cite{paliakaite2021modeling} caused by sensor displacement, forearm and hand motion, and poor contact, leading to missed episodes and false alarms. Additionally, irregular rhythms such as premature atrial contractions, bigeminy, atrial flutter, and tachycardia contribute to false positives~\cite{petrenas2015low,solovsenko2019detection}. Therefore, further research is needed to determine whether machine learning algorithms trained on raw PPG data or those incorporating feature extraction and signal quality evaluation are more effective in reducing uncertainty in AF detection.\\

The development of reliable AF detectors using PPG is hindered by the limited availability of high-quality labeled datasets. The lack of guidelines for arrhythmia interpretation in PPG necessitates alternative annotation strategies such as the simultaneous acquisition of ECG for AF verification. As a result, existing PPG datasets are either too small or contain inaccurate labels due to reliance on automated annotations rather than expert verification.

\subsection{Potential datasets}
\label{sec:datasets_af}
AF PPG datasets are categorized into wrist-based and fingertip-based datasets for training and testing purposes.\\

\textit{Wrist-based datasets:}\\

\textbf{DeepBeat} is the preferred dataset for training due to its large scale and existing code support, despite some label noise. The dataset comprises over 3336185 25-second PPG segments sampled at 32 Hz from 167 individuals. PPG signals were collected using a wrist-based wearable device in various conditions, including before cardioversion, during an exercise stress test, and in daily life. The dataset was restructured in \cite{D1} to ensure an equal AF/non-AF ratio across training, validation, calibration, and test sets eliminating redundancy from the original data split and providing a more reliable representation for model evaluation. The final distribution includes the following splitting pattern:\\
\begin{itemize}
\item Training set: 40253 AF samples from 44 subjects and 65489 non-AF samples from 54 subjects.
\item Validation set: 5749 AF samples from 20 subjects and 9343 non-AF samples from 16 subjects.
\item Calibration set 5753 AF samples from 20 subjects and 9392 non-AF samples from 15 subjects.
\item Test set: 5746 AF samples from 19 subjects and 9,343 non-AF samples from 17 subjects.
\end{itemize}
The code to generate this final distribution is described in deliverable D1 and \cite{D1}.\\

\begin{figure}[ht]
    \centering
    \includegraphics[width=12cm]{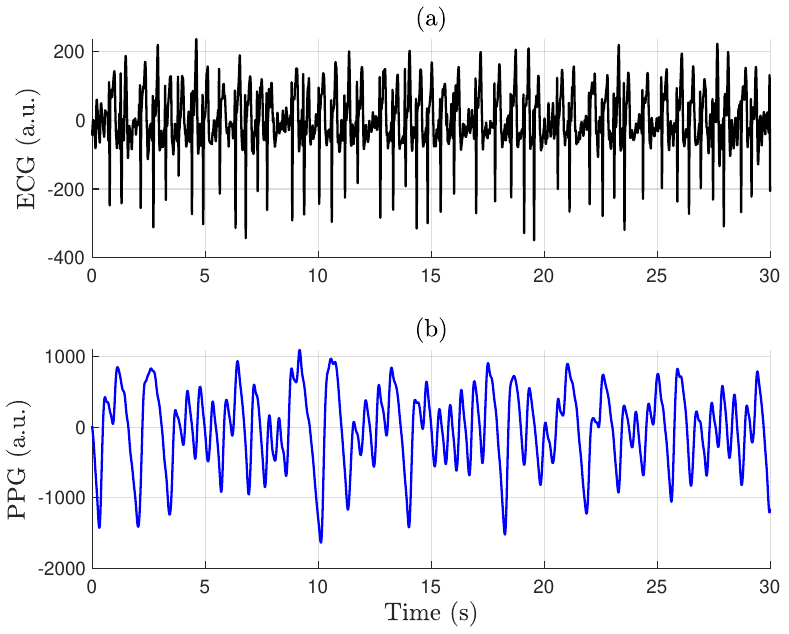}
    \caption{The example of (a) ECG segment and (b) PPG segment with atrial fibrillation from TriggersAF dataset.}
    \label{ECG_PPG_TriggersAF}
\end{figure}

\begin{figure}[ht]
    \centering
    \includegraphics[width=12cm]{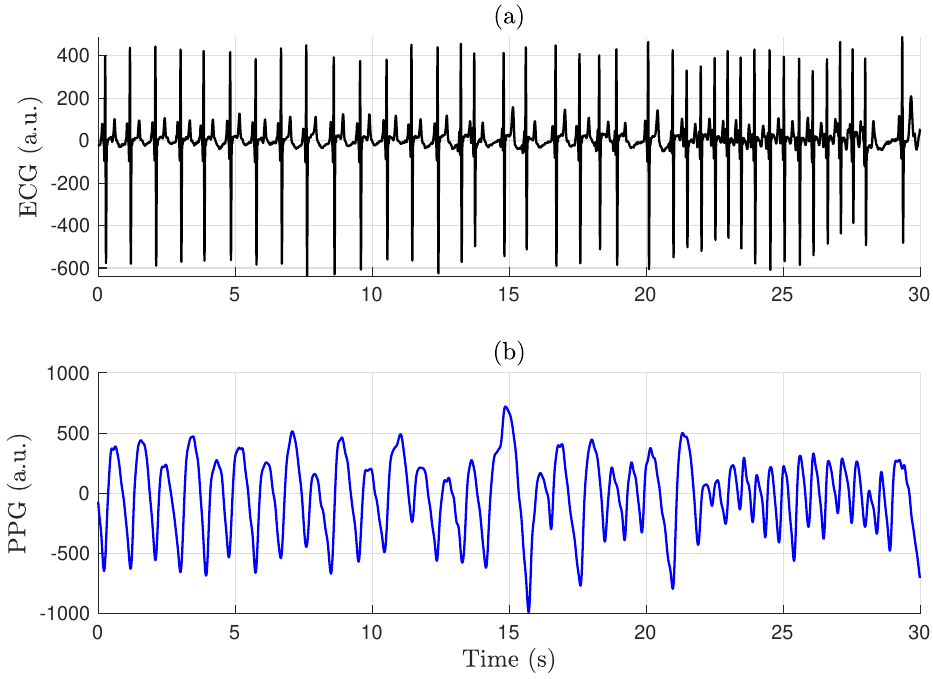}
    \caption{The example of (a) ECG segment and (b) PPG segment with premature beats and tachycardia from TriggersAF dataset.}
    \label{ECG_PPG_TriggersAF_tachy}
\end{figure}

The \textbf{TriggersAF} dataset \cite{Bacevicius2024} includes data from patients diagnosed with paroxysmal AF, recruited from inpatient and outpatient wards of the Cardiology Department at Vilnius University Hospital Santaros Klinikos. Prior to participation, all patients provided written informed consent in accordance with the ethical principles of the Declaration of Helsinki, and the study received approval from the Vilnius Regional Bioethics Committee (Reference Number 158200-18/7-1052-557). The dataset comprises 133 subjects (48 female, 85 male), with an average age of 57.9$\pm$11.6 years and a BMI of 28.4$\pm$4.9. There are no cases of permanent AF, with 45 subjects experiencing AF and 88 classified as non-AF. The average duration of recordings is 6.9$\pm$1.1 days, capturing a total of 307 AF episodes, with an individual average of 10$\pm$19 episodes. The AF burden is approximately 6.8$\pm$26\%. ECG and PPG signals are sampled at 500 Hz and 100 Hz, respectively. Given its quality, TriggersAF is being evaluated for use as either a training set (whole samples) or a high-quality test set. An example of synchronized ECG and PPG signals from the TriggersAF dataset with atrial fibrillation as well as premature beats and tachycardia are shown in figures~\ref{ECG_PPG_TriggersAF} and~\ref{ECG_PPG_TriggersAF_tachy}, respectively.\\

The \textbf{UMMC Simband} dataset \cite{Bashar2019, Han2020} consists of 37 patients (28 male, 9 female) aged between 50 and 91 years, all diagnosed with cardiac arrhythmia. Participants wore the Samsung Simband 2 smartwatch, and their ECG was taken using a 7-lead Holter monitor. The data were preprocessed and segmented into 30-second windows with no overlap. The ECG signals were sampled at 128 Hz, while PPG signals were downsampled to 50 Hz. The signals were labeled into five categories: 0 (normal sinus rhythm), 1 (atrial fibrillation), 2 (premature atrial contractions/premature ventricular contractions), 3 (uncertain if NSR or PAC/PVC), and 5 (noisy PPG). For this study, only the PPG signals were used, and they were labeled as AF (1) or normal (0, 2, 3). Labels 5 and NaN (no ECG reference) were excluded from the analysis. The Simband dataset was discussed as a test set, providing additional validation for wrist-based AF detection models.\\

\textit{Fingertip-based datasets:}\\

\textbf{MIMIC-III-Ext-PPG} \cite{PhysioNet-mimic-iii-ext-ppg-1.1.0, Goldberger2000} is a new PPG-based benchmark dataset that has been created during the course of the QUMPHY project. It is a large-scale dataset based on the MIMIC III Waveform Database capturing almost five million 30s PPG waveform segments from more than 6000 patients. Rhythm annotations were inferred from heart rhythm chart events in the corresponding MIMIC-III clinical database \cite{johnson2016mimic}. It covers 26 different rhythm types, including atrial fibrillation and atrial flutter, which allows to use the dataset as a benchmark dataset for robust atrial fibrillation detection, using all other rhythm classes as negative examples. It is particularly suited for training due to its size and diverse rhythm types. A potential second MIMIC-III-based dataset \cite{BasharIEEE} with manual AF annotations was used for technical validation of the label quality of the MIMIC-III-Ext-PPG dataset and is therefore not considered as a benchmark dataset. 

\textbf{Liu2022} The way of rhythm annotations were derived in MIMIC-III-Ext-PPG necessitates a second fingertip-based dataset for external validation. Here, we leverage the dataset prepared by Liu et al \cite{Liu2022}, which provides 91 10s PPG segments from 91 patients annotated according to five different rhythm types (sinus rhythm, premature ventricular contraction, premature atrial contraction, ventricular tachycardia, supraventricular tachycardia, atrial fibrillation). The dataset is not derived from MIMIC-III and therefore ideally suited as an external validation dataset for MIMIC-III-Ext-PPG.

\subsection{Dataset selection}
We recommend to use one large training set for wrist based measurements and one for fingertip measurements together with an additional test set in both situations. Concerning wrist based measurements, we chose DeepBeat as training set and TriggersAF as test set. For the fingertip measurements, we take MIMIC PPG as training set and the dataset from Liu as external test set.

\subsection{Making the datasets available}
The qumphy version of DeepBeat has already been described in the D1 report and \cite{D1}. Triggers AF is available on zenodo \cite{Bacevicius2024}. 
The MIMIC-III-Ext-PPG dataset \cite{PhysioNet-mimic-iii-ext-ppg-1.1.0} has been published recently on Physionet \cite{Goldberger2000}. %
The Liu2022 dataset is publicly available from an associated github repository \cite{liu2024github}. The dataset is supposed to be used in its entirety as a test set.

\subsection{Dataset usage}

\textit{Wrist-based datasets:}

In the previous report D1 of the qumphy project and \cite{D1}, it was already described how to train models on DeepBeat with the following splitting pattern:\\
Training set: 40253 AF samples from 44 subjects and 65489 non-AF samples from 54 subjects.\\
Validation set: 5749 AF samples from 20 subjects and 9343 non-AF samples from 16 subjects.\\
Calibration set: 5753 AF samples from 20 subjects and 9392 non-AF samples from 15 subjects.\\
Test set: 5746 AF samples from 19 subjects and 9,343 non-AF samples from 17 subjects.
This is supposed to be complemented by TriggersAF as external test set.

\textit{Fingertip-based datasets:}

MIMIC-III-Ext-PPG comes with 13 predefined stratified splits that can be used for structured benchmarking. In particular, the final fold is supposed to be used as internal test set. The Liu2022 dataset will be used as external test set to assess the generalizability of models trained on MIMIC-III-Ext-PPG.

\section{Benchmark III: Classification of hypertension}%
\label{sec:benchmark_3}

\subsection{Problem}
Elevated blood pressure and hypertension has a very high prevalence, i.e. around 30\% of people in Europe are affected. It is one of the major risk factors for many diseases, in particular cardiovascular diseases. This includes Hypertension-mediated organ damage and hypertension is associated with cardiovascular and non-cardiovascular outcomes such as stroke, cognitive impairment, heart failure, atrial fibrillation or diabetes \cite{ESCguidelines}.

\subsection{Potential datasets}
\subsubsection{AURORABP}
As presented in Section \ref{AURORABP_dataset_presentation_section}, the \textbf{AuroraBP} dataset \cite{auroraBP} consists of PPG, ECG and BP signals for two separate cohorts. What differentiates the two cohorts is the technique used for blood pressure measurement (auscultatory or oscillometric) and the presence or absence of ambulatory measurements. The study comprised of 1,125 participants aged 21-85, of which 49.2\% were female.

\subsection{Making the datasets available}
The AURORABP dataset is available upon request and approval from the data regulatory committee, as described in Section \ref{AURORABP_access_section}. \\
The MatLab code that has been used to split the data into different classes and to preprocess the data will be made publicly available, to allow for reproducibility.  
\subsection{Dataset usage}
Multiple records were present for each subject, and they are contained in a single fold. The folds have been stratified to give similar distributions of the following classes:

\begin{itemize}
\item 
Gender: Male/female.
\item 
Blood pressure class: Three classes are defined in terms of systolic blood pressure (SBP) and diastolic blood pressure (DBP), and in accordance with the 2024 ESC Guidelines \cite{ESCguidelines} by
\begin{itemize}
\item 
Non-elevated: $\mathrm{SBP}< 120$ mmHg and $\mathrm{DBP}< 70$ mmHg
\item 
Elevated: $120\leq\mathrm{SBP}< 140$ mmHg or $70\leq\mathrm{DBP}< 90$ mmHg
\item 
Hypertensive: $\mathrm{SBP}\geq 140$ mmHg or $\mathrm{DBP}\geq 90$ mmHg
\end{itemize}
Note that the Aurora BP variables \emph{baseline\_sbp} and \emph{baseline\_dbp} are used for the SBP and DBP values \cite{auroragitreference}. These provide a single value for each subject, which then allowed classification into one of the 3 classes.
\item 
Cardiovascular disease: The Aurora BP metadata contains information about a range of self-reported cardiovascular diseases. Two classes are used, namely no cardiovascular disease and cardiovascular disease if at least one condition is reported.
\item 
Body mass index (BMI): Three classes of BMI are used:
\begin{itemize}
\item 
Healthy: $\mathrm{BMI}<25$ kg/m$^2$
\item
Overweight: $25\leq\mathrm{BMI}<30$ kg/m$^2$
\item
Obese: $\mathrm{BMI}\geq 30$ kg/m$^2$
\end{itemize}
\end{itemize}

\begin{table}[H]
\caption{Population characteristics for the BP classification and regression cohort - HW: Healthy Weight, OW: Overweight, OB: Obese. Age, weight and height are expressed as mean $\pm$ standard deviation.}
\renewcommand\arraystretch{1.3}
    \begin{tabular}{|c | *{5}{>{\centering}p{2cm}|}c|}
\hline
\multicolumn{7}{|c|}{} \\
 \multicolumn{7}{|c|}{\textbf{Total: 1100}: 557M, 543F (49.4\%F)}\\
\multicolumn{7}{|c|}{} \\
 \hline
    & \multicolumn{2}{c|}{\textbf{Non Elevated}}  & \multicolumn{2}{c|}{\textbf{Elevated}}& \multicolumn{2}{c|}{\textbf{Hypertensive}}\\
     & \multicolumn{2}{c|}{\textbf{213 (19.4\%)}}  & \multicolumn{2}{c|}{\textbf{645 (58.6\%)}}& \multicolumn{2}{c|}{\textbf{241 (21.9\%)}}\\
      & \multicolumn{2}{c|}{73M / 140 F (65.7\%F)}  & \multicolumn{2}{c|}{337M / 308F (47.8\%F)}& \multicolumn{2}{c|}{146M / 95F (39.4\%F)}\\
    \hline
\textbf{CVD} & \textbf{no cvd} & \textbf{cvd} &\textbf{no cvd} & \textbf{cvd} &\textbf{no cvd} & \textbf{cvd} \\
 & 172 (76.5\%) & 53 (23.5\%) & 374 (59.0\%) & 260 (41.0\%) & 77 (32.0\%) & 164 (68\%) \\
 \hline
 \textbf{Gender} & 51 / 114  & 22 / 26 & 189 / 191 & 148 / 117 & 53 / 24  & 93 / 71 \\ 
 (M/F) &(69.1\% F) & (54.2\% F) & (50.2\% F) & (44.2\% F) & (31.2\% F) & (43.3\% F)\\
 \hline 
\textbf{BMI} & HW & HW & HW & HW & HW & HW \\
& 90 (54.6\%) & 12 (25.0\%) & 128 (33.7\%) &  39 (14.7\%) & 18 (23.4\%) & 21 (12.8\%) \\  
 & OW & OW & OW & OW & OW & OW \\
&46 (27.9\%) & 14 (29.2\%) & 143 (37.6\%) & 94 (35.4\%) & 27 (35.1\%) & 52 (31.7\%) \\
 & OB & OB & OB & OB & OB & OB \\
&29 (17.6\%) & 22 (45.8\%) & 109 (28.7\%) & 132 (49.8\%) & 32 (41.5\%) & 90 (54.9\%) \\ 
&&&&&& (1 NaN) \\
\hline 
\textbf{Age} & 38.2$\pm$10.28 & 45.81$\pm$11.65 & 43.06$\pm$10.31 & 48.46$\pm$10.39 & 46.85$\pm$12.47 & 51.35$\pm$10.58 \\
\hline 
\textbf{Height (m)} & 1.68$\pm$0.09 & 1.71$\pm$0.11&1.71$\pm$0.10 & 1.72$\pm$0.10 & 1.75$\pm$0.09&1.71$\pm$0.10\\
\hline 
\textbf{Weight (Kg)}&73.56$\pm$18.38 & 93.18$\pm$34.08 & 82.67$\pm$19.80 & 93.16$\pm$23.60 & 91.34$\pm$22.37&93.36$\pm$21.84\\
\hline 
\end{tabular}

\label{BP population characteristics table}
    \end{table}

This large collection of data allows you to perform several versions of classification tasks between non-elevated, elevated and hypertensive patients and to analyse the influence of many risk factors. Matlab code that has been used to create stratified 10-fold splits for these tasks will be made publicly available.

\section{Benchmark IV: Classification / regression vascular age}%
\label{sec:benchmark_4}

\subsection{Problem}
Vascular Ageing (VA) is a complex process that involves the gradual deterioration of arterial structure and function over time, negatively impacting organ function \cite{climie2023vascular}. The gold-standard measurement for VA is carotid-femoral pulse wave velocity, which requires trained personnel and is not routinely clinically available \cite{reshetnik2017oscillometric}. In healthy ageing, chronological and vascular age typically correspond \cite{hamczyk2020biological}.\\
Early detection of premature VA is critical for the timely identification and treatment of cardiovascular disease (CVD), which remains a leading global health burden.\\
Non-invasive signals from photoplethysmography (PPG) or tonometry can help assess vascular age, by analysing the shape of the pulse wave, which changes with age due to arterial stiffening. PPG is an optical method used in clinical and wearable devices to measure pulse waves at sites like the wrist and finger \cite{charlton2022}. Arterial tonometry, mainly used clinically, measures pressure from superficial arteries such as the radial or carotid \cite{salvi2015noninvasive}. 
By comparing signal-based estimates of vascular age to a person's chronological age, we hypothesised we could identify early VA in community based settings.

\subsection{Potential datasets}
\subsubsection{AURORABP}
As presented in Section \ref{AURORABP_dataset_presentation_section}, the \textbf{AuroraBP} dataset \cite{auroraBP} consists of PPG, ECG and BP signals for two separate cohorts. What differentiates the two cohorts is the technique used for blood pressure measurement (auscultatory or oscillometric) and the presence or absence of ambulatory measurements. The study comprised of 1,125 participants aged 21-85, of which 49.2\% were female.

\subsubsection{Pulse Wave Database (PWDB)}
Another potential dataset is the Pulse Wave Database \cite{charlton2019}, a dataset of simulated waveforms that comprises of 4,374 healthy male subjects divided into six 10-year age groups, spanning from 25 to 75 years of age (25, 35, 45, 55, 65, 75). It contains single-cycle PPG signals, among others, at various locations in the body and is free to download and use.

\subsection{Dataset selection}
The AURORABP dataset was used for this task. Participants were separated into "suitable" and "unsuitable". Participants who were marked as "suitable" satisfied \textbf{all} of the following conditions: no cvd, no hypertension, and no obesity.\\
Participants were labeled as \textbf{hypertensive} if SBP was over 140 mmHg or DBP was greater than 90 mmHg, in agreement with the 2024 ESC guidelines \cite{2024_esc_hypertension_guidelines}. They were labeled as \textbf{obese} if their BMI was $\geq$ 30, while they were assigned to the "CVD" (cardiovascular disease) if their records contained at least one of the following: 
\begin{itemize}
    \item  high blood pressure (not highly reliable according to published study)
    \item  coronary artery disease
    \item  diabetes
    \item  arrythmia
    \item  previous heart attack
    \item  previous stroke
    \item  heart failure
    \item  aortic stenosis
    \item  valvular heart disease
    \item  "other cv diseases" (not better specified)
    \item  cvd meds (not better specified)
\end{itemize}

\noindent Participants were then separated into 10-year age groups: 
\begin{itemize}
    \item $<$ 30
    \item 30-40
    \item 40-50
    \item 50-60
    \item 60-70
    \item 70+,
       
\end{itemize}
and an overview of the population composition is presented in Table \ref{Age population characteristics table}.

\noindent \textbf{Note}: separation by gender was not accounted for because the gender distribution was uneven across ages, even though it might impact ppg waveforms. The limitations mentioned in the BP section remain valid also for the Age regression section. 
\begin{table}[H]
\caption{Population characteristics for the age regression cohort - HW: Healthy Weight, OW: Overweight, OB: Obese. Age, weight, height, SBP and DBP are expressed as mean $\pm$ standard deviation. SBP and SBP values referred here are the "baseline\_sbp" and "baseline\_dbp" values used to divide the population into BP classes.}
\renewcommand\arraystretch{1.3}
    \begin{tabular}{|c | *{5}{>{\centering}p{2cm}|}c|}
\hline
\multicolumn{7}{|c|}{} \\
 \multicolumn{7}{|c|}{\textbf{Total: 546}}\\
\multicolumn{7}{|c|}{} \\
 \hline
 & \textbf{$<$30}  & \textbf{30-39}& \textbf{40-49}&\textbf{50-59}&\textbf{60-69}&\textbf{70+}\\  
\textbf{Age group}    & 57 & 205 & 146 & 118 & 16 & 4 \\
    & (10.4\%) & (37.5\%)&(26.7\%)&(21.7\%)&(3.0\%)&(0.7\%)\\
    \hline
\textbf{Height (m)} & 1.70$\pm$0.10 & 1.71$\pm$0.10&1.71$\pm$0.10&1.70$\pm$0.09&1.69$\pm$0.09&1.72$\pm$0.10\\
\hline 
\textbf{Weight (Kg)}&70.41$\pm$15.21&82.49$\pm$21.44&81.59$\pm$19.65&79.42$\pm$18.65&70.32$\pm$9.28&75.40$\pm$16.39\\
\hline 
\textbf{SBP}& 114.16$\pm$10.55&117.34$\pm$10.53&118.38$\pm$10.11&118.90$\pm$10.49&125.49$\pm$11.66&131.66$\pm$5.68\\
\hline
\textbf{DBP}&67.03$\pm$8.80&71.82$\pm$7.94&73.62$\pm$8.22&74.25$\pm$6.86&72.63$\pm$7.46&67.01$\pm$4.46\\
\hline
\end{tabular}

\label{Age population characteristics table}
    \end{table}

\subsection{Making the datasets available}
The AURORABP dataset is available upon request and approval from the data regulatory committee, as described in Section \ref{AURORABP_access_section}. \\
The MatLab code that has been used to split the data into different classes and to preprocess the data will be made publicly available, to allow for reproducibility.  \\
The PWDB dataset is free to download on Zenodo \cite{pwdbzenodo}.
\subsubsection{Dataset usage}
The population was split into 10 separate folds, balanced for age and sex, that can then be used for train, validation, calibration and test. 

\section{Benchmark V: Detection of sleep apnea}%
\label{sec:benchmark_5}
\subsection{The problem}

Obstructive sleep apnea (OSA) is a prevalent sleep disorder characterized by repeated episodes of partial or complete upper airway obstruction during sleep, leading to disrupted breathing, oxygen desaturation, and fragmented sleep \cite{comprehensive_abbasi_2021, clinical_kapur_2017}. Its clinical significance is significant, as untreated OSA is associated with a range of comorbidities, including hypertension, cardiovascular disease, stroke, type 2 diabetes, and neurocognitive impairments such as memory deficits and impaired executive function \cite{obstructive_yeghiazarians_2021, obstructive_bubu_2020}. In addition, OSA contributes to daytime fatigue, increasing the risk of car accidents and workplace errors \cite{sleep_smolensky_2011}. The chronic physiological stress from recurrent apneas and hypopneas triggers systemic inflammation and oxidative stress, further exacerbating metabolic and cardiovascular complications. The condition’s impact on quality of life, coupled with its economic burden due to healthcare costs and lost productivity, underscores the need for timely diagnosis and effective management \cite{economic_nieden_2024}.\\

Detecting OSA poses significant challenges, primarily due to underdiagnosis and limited access to diagnostic tools. Polysomnography (PSG), the gold standard for diagnosis, is resource-intensive, requiring overnight monitoring in a sleep laboratory with specialized equipment and trained personnel \cite{clinicianfocused_markun_2020}. This makes it costly and inaccessible for many patients, particularly in underserved or rural areas \cite{polysomnography_hirshkowitz_2016}. Home sleep apnea testing (HSAT) offers a more accessible alternative, but its sensitivity and specificity vary, and it may miss milder cases or misclassify complex sleep disorders \cite{clinical_rosen_2017}. Moreover, many individuals with OSA remain asymptomatic or attribute symptoms like snoring or daytime sleepiness to other causes, delaying presentation to healthcare providers. Public awareness of OSA is low, and screening tools, such as questionnaires (e.g., STOP-Bang), while useful, lack precision and can lead to over- or under-referral for testing \cite{stopbang_costa_2020, screening_evans_2017}. These detection barriers highlight the need for scalable, cost-effective diagnostic solutions, such as wearable devices and AI-driven screening tools, to improve early identification and treatment of OSA.\\

\begin{figure}[ht] 
	\centering
	\includegraphics[width=15.5cm]{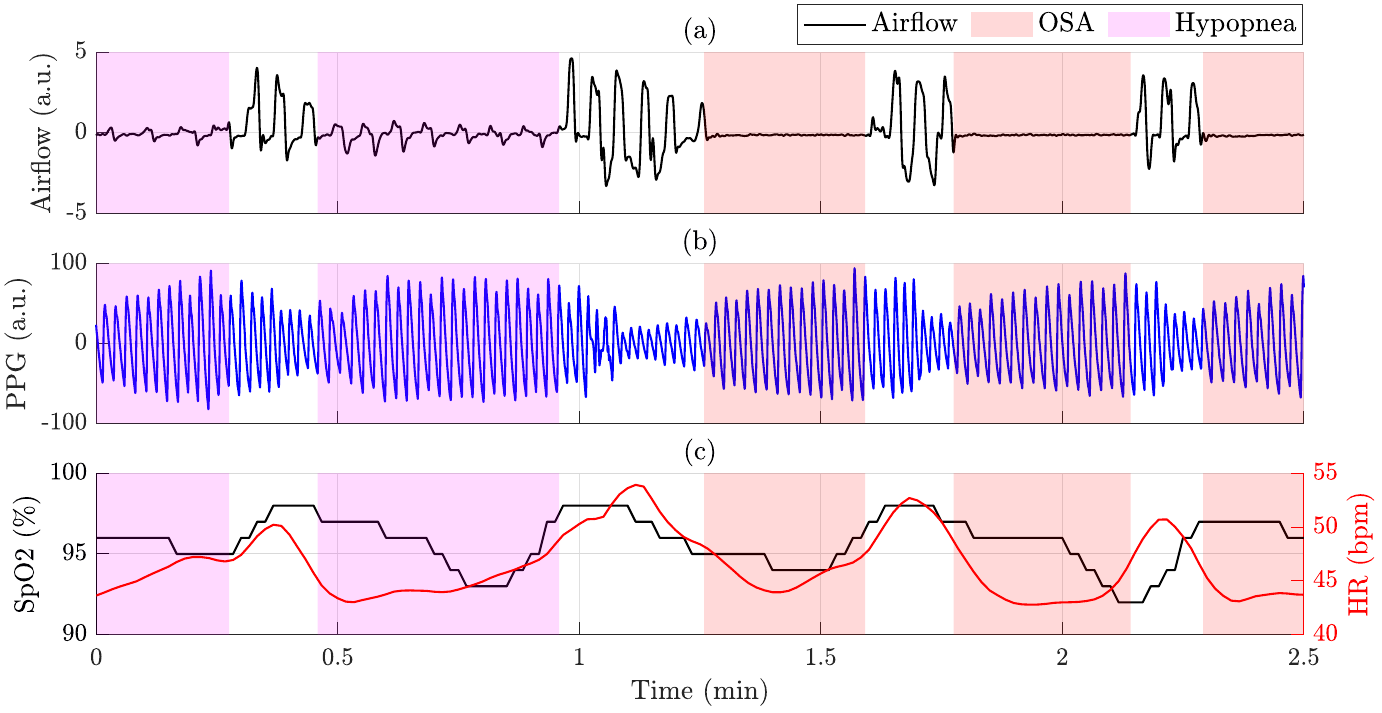} 
	\caption{The changes of signals during sleep apnea episodes: (a) respiratory airflow, (b) finger PPG signal, (c) arterial blood oxygen saturation, SpO2, and heart rate, HR.}
	\label{PPG_Apnea_SpO2_HR}
\end{figure}

An apnea event is characterized as a lack of airflow for at least 10~s with an associated oxygen desaturation of at least 3-4\% and/or arousals. A hypopnea event is characterized as a 50\% or greater reduction in airflow for at least 10~s with an associated oxygen desaturation of at least 3-4\% and/or arousal. Episodes of apneas and hypopneas in the airflow, PPG, heart rate (HR) and SpO2 signals are illustrated in the Figure \ref{PPG_Apnea_SpO2_HR}.

\newpage

\subsection{Potential datasets}

\subsubsection{OSASUD}

The open-access \textbf{OSASUD} polysomnography (PSG) dataset~\cite{OSASUD} involves 30 patients (36.67\% of female) admitted to the stroke unit at the Clinical Neurology Unit of Udine University Hospital between 2019-2020 due to suspected cerebrovascular events, such as ischemic and hemorrhagic stroke, or transient ischemic attack. The dataset provides information of subject age (68.97 $\pm$ 11.22 years), the body-mass index (29.17 $\pm$ 5.74 $kg/m^{2}$), the apnea-hypopnea index (25.43 $\pm$ 21.31 counts/h), and signal duration (8.91 $\pm$ 1.47 h). Patients of age $<$18 years, those unable to comply with standard  requirements for PSG monitoring, individuals with severe aphasia impacting understanding or consent, and those at high risk of alcohol or drug withdrawal syndrome were excluded from the study. Notably, conditions such as obesity, diabetes mellitus, atrial fibrillation and other cardiac disease did not serve as exclusion criteria.\\

The dataset includes PSG signals such as finger photoplethysmography (PPG), electrocardiography (ECG), oxygen saturation (SpO2), respiratory rate, perfusion index, heart rate, nasal airflow, snoring, 3-axis accelerometer, abdominal and thoracic movements. The sampling rate of PPG and ECG signals is 80 Hz. While respiratory rate and SpO2 time series are sampled at 1 Hz. In addition, SpO2 values $<$50\% or $>$100\% were identified as artifacts and marked as null. This method was similarly applied to heart rate measurements $<$20 or $>$200, as well as respiratory rate values $<$5 or $>$40.\\

Recorded PSG data underwent thorough analysis using Embla RemLogic software, version 3.4.1.2371 (Natus Medical Inc., Pleasanton, CA, USA), which facilitates signal processing, detailed examination, and annotation procedure. The OSASUD data were annotated by a trained sleep medicine physician based on the sleep scoring guidelines established by the American Academy of Sleep Medicine, identifying occurrences of central, obstructive, mixed apnea, and hypopnea events (1~s~granularity). The OSASUD dataset has already been used in the following state-of-the-art studies~\cite{OSASUD2, OSASUD3, OSASUD4, OSASUD5}.

\subsubsection{MESA}\label{MESA_dataset_presentation_section}

The \textbf{MESA} dataset~\cite{MESA} comprises data from 2055 patients (53.63\% of female), aged between 54 and 95 years. It contains a total of 16~300 hours of fully annotated overnight PSG recordings collected during a sleep study (2010-2012), funded by the National Heart, Lung, and Blood Institute. The average age of subjects is 69.37 $\pm$ 9.12 years. Participants included in the study had not used treatments for sleep apnea, such as continuous positive airway pressure or oxygen devices, for more than a month prior to the study, or only used it less than once a week. The dataset represents a cohort, featuring participants from a variety of racial/ethnic groups, including White, African American, Hispanic, and Chinese American individuals.\\

PSG recordings were obtained at patient home using the Compumedics Somte monitoring system (Compumedics Ltd., Abbotsville, Australia). The dataset includes signals such as PPG, ECG, electrooculography-EOG, electroencephalography-EEG, electromyography-EMG, SpO2, nasal airflow, snoring, abdominal and thoracic movements. The sampling rate of PPG and ECG signals is 256 Hz. PPG signals were recorded from the finger using the Nonin 8000 sensor. In addition, time labels and durations of apnea and hypopnea episodes in respiratory flow signal are annotated. However, this database is not open-access and is only available on request. The MESA dataset has already been used in the following state-of-the-art studies~\cite{MESA2, MESA3}.

\subsection{Making the datasets available}

The OSASUD is fully open access, whereas the MESA is only available on request. The original OSASUD dataset can be downloaded from \href{https://springernature.figshare.com/collections/A_dataset_of_stroke_unit_recordings_for_the_detection_of_Obstructive_Sleep_Apnea_Syndrome/5630890}{here}. The MESA dataset is not freely available, and an application for access to the data should be made to the National Sleep Research Resource. This can be done \href{https://sleepdata.org/datasets/mesa}{here}. The Matlab codes to extract 10 folds/splits from OSASUD and MESA datasets will be available at QUMPHY \href{https://gitlab.com/qumphy/d4-code/}{repository} very soon.~While the OSASUD is freely available dataset, *.mat and *.pkl derived files are additionally provided.

\subsection{Dataset usage}

In the following, an introduction to how to use these datasets for the problem of sleep apnea detection from PPG signals is given.

\subsubsection{OSASUD}

The reduced OSASUD dataset file (\emph{OSASUD\_initial}), the Matlab code (\emph{Data\_Segmentation\_OSASUD}) which splits this OSASUD file into 10 folds, and these folds will be published very soon. The training, validation, calibration, and test sets can be defined by the user from these 10 folds as appropriate, with a split of 7/1/1/1 folds being suggested respectively (or 8/1/1 or 7/2/1 if a calibration set is not required). Using the folds for cross-validation is clearly also an option. All the records for each subject are contained in a single fold. These folds have been stratified to give similar distributions of apnea-hypopnea index (AHI) values.\\

The OSASUD consists of 30 subjects, so each fold consists of nocturnal raw PPG signals and estimated features from 3 subjects. Segment duration - 1 min (raw PPG segment - 4800 samples, PPG feature sequence - 60 samples).\\

\emph{Data\_Segmentation\_OSASUD} code processes:

\begin{enumerate}
\item \emph{OSASUD\_initial} file in order to obtain \emph{OSASUD\_segments} file with 1-min raw PPG intervals and PPG feature sequences for model training/validation/calibration/testing. 
\item \emph{OSASUD\_segments} file in order to obtain \emph{OSASUD\_PPG\_feature\_folds}, \emph{OSASUD\_PPG\_segment\_folds}, \emph{OSASUD\_Apnea\_label\_folds}, and \emph{OSASUD\_RRate\_label\_folds} files with 10 folds (10 groups of 3 subjects) stratified according to AHI labels.
\end{enumerate}

\emph{OSASUD\_initial} file includes data of the table, whose columns are:

\begin{enumerate}
\item Subject ID;
\item SpO2 time series (sampling rate - 1 Hz);
\item Apnea event type (APNEA-OBSTRUCTIVE, HYPOPNEA, APNEA-CENTRAL, APNEA-MIXED, ALL APNEAS);   
\item Anomaly (1 - if there is any apnea event, 0 - otherwise);
\item PPG signal (sampling rate - 80 Hz);
\item Respiratory rate (sampling rate - 1 Hz).
\end{enumerate}

\emph{OSASUD\_segments} file includes estimated PPG features, raw PPG segments, and labels divided for 1-min 13,829 PPG intervals:

\begin{enumerate}
\item \emph{X\_cell\_features} includes 4 features (1 - pulse interval, 2 - peak-to-peak amplitude, 3 - area-related feature, and 4 - SpO2) sampled at 1 Hz (segment duration - 60 samples) for 30 subjects.
\item \emph{X\_cell\_ppg\_segments} includes raw 1-min PPG segments sampled at 80 Hz (segment duration - 4800 samples) for 30 subjects.
\item \emph{Y\_cell} includes 6 category labels for 30 subjects:

\begin{enumerate}
\item APNEA-OBSTRUCTIVE events;
\item HYPOPNEA events;
\item APNEA-OBSTRUCTIVE \& HYPOPNEA events;
\item APNEA-CENTRAL events;
\item APNEA-MIXED events;
\item ALL APNEAS.
\end{enumerate}

\item \emph{Y\_cell\_rr} includes mean respiratory rate labels for 30 subjects.

\end{enumerate}

\emph{OSASUD\_PPG\_feature\_folds} file includes 10 folds (10 groups of 3 subjects) of estimated PPG features (1 - pulse interval, 2 - peak-to-peak amplitude, 3 - area-related feature, and 4 - SpO2). PPG feature sequence duration - 60 samples (1-min).
 
\emph{OSASUD\_PPG\_segment\_folds} file includes 10 folds (10 groups of 3 subjects) of raw PPG segments. PPG segment duration - 4800 samples (1-min).

\emph{OSASUD\_Apnea\_label\_folds} file includes 10 folds (10 groups of 3 subjects) of apnea labels (1 - APNEA-OBSTRUCTIVE events, 2 - HYPOPNEA events, 3 - APNEA-OBSTRUCTIVE \& HYPOPNEA events, 4 - APNEA-CENTRAL events, 5 - APNEA-MIXED events, 6 - ALL APNEAS) for 1-min PPG signals.

\emph{OSASUD\_RRate\_label\_folds} file includes 10 folds (10 groups of 3 subjects) of respiratory rate labels for 1-min PPG signals.\\

Note 1: we suggest using '3 - APNEA-OBSTRUCTIVE \& HYPOPNEA events' label for detecting apnea segments.

Note 2: instead of using raw PPG segments, we suggest using estimated (a) 3 PPG features (pulse interval, peak-to-peak amplitude, area-related feature) and (b) SpO2 time series as model inputs, separately, and then compare the (a) and (b) results.

Note 3: all data - 30 subjects, 13,829 1-min segments.

\subsubsection{MESA}

The Matlab code (\emph{Data\_Segmentation\_MESA}) which splits the MESA dataset into 10 folds will be published very soon. The training, validation, calibration, and test sets can be defined by the user from these 10 folds as appropriate, with a split of 7/1/1/1 folds being suggested respectively (or 8/1/1 or 7/2/1 if a calibration set is not required). Using the folds for cross-validation is clearly also an option. All the records for each subject are contained in a single fold. These folds have been stratified to give similar distributions of AHI values.\\

The original MESA consists of 2055 subjects. For signal quality and class balance issues, we used only 160 subjects with 'PSG Study Quality Grade = 7' (highest quality) and 'AHI $>$ 25'. Thus, each fold consists of nocturnal raw PPG signals and estimated features from 16 subjects. Segment duration - 1 min (raw PPG segment - 15360 samples, PPG feature sequence - 60 samples).\\

\emph{Data\_Segmentation\_MESA} code processes:

\begin{enumerate}
\item 2055 *.edf PPG signals in order to obtain \emph{MESA\_segments} file with 1-min raw PPG intervals and PPG feature sequences for model training/validation/calibration/testing, and \emph{MESA\_metadata} file with subject AHI values. For signal quality and class balance issues, \emph{MESA\_segments} and \emph{MESA\_metadata} files include information of only 160 subjects with 'PSG Study Quality Grade = 7' (highest quality) and 'AHI $>$ 25'.
\item \emph{MESA\_segments} file in order to obtain \emph{MESA\_PPG\_feature\_folds}, \emph{MESA\_PPG\_segment\_folds}, and \emph{MESA \_Apnea\_label\_folds} files with 10 folds (10 groups of 16 subjects) stratified according to AHI labels.
\end{enumerate}

\emph{MESA\_segments} file includes estimated PPG features, raw PPG segments, and labels divided for 1-min PPG intervals:

\begin{enumerate}
\item \emph{X\_cell\_features} includes 4 features (1 - pulse interval, 2 - peak-to-peak amplitude, 3 - area-related feature, and 4 - SpO2) sampled at 1 Hz (segment duration - 60 samples) for 160 subjects.
\item \emph{X\_cell\_ppg\_segments} includes raw 1-min PPG segments sampled at 256 Hz (segment duration - 15360 samples) for 160 subjects.
\item \emph{Y\_cell} includes 3 category labels for 160 subjects:

\begin{enumerate}
\item APNEA-OBSTRUCTIVE events;
\item HYPOPNEA events;
\item APNEA-OBSTRUCTIVE \& HYPOPNEA events.
\end{enumerate}

\end{enumerate}

\emph{MESA\_PPG\_feature\_folds} file includes 10 folds (10 groups of 16 subjects) of estimated PPG features (1 - pulse interval, 2 - peak-to-peak amplitude, 3 - area-related feature, and 4 - SpO2). PPG feature sequence duration - 60 samples (1-min).
 
\emph{MESA\_PPG\_segment\_folds} file includes 10 folds (10 groups of 16 subjects) of raw PPG segments. PPG segment duration - 15360 samples (1-min).

\emph{MESA\_Apnea\_label\_folds} file includes 10 folds (10 groups of 16 subjects) of apnea labels (1 - APNEA-OBSTRUCTIVE events, 2 - HYPOPNEA events, 3 - APNEA-OBSTRUCTIVE \& HYPOPNEA events) for 1-min PPG signals.\\

\section{Benchmark VI: Regression respiratory rate}%
\label{sec:benchmark_6}

 \subsection{Problem}

 Respiratory rate (RR) gives much information of the physiological state \cite{Braun1990, Charlton2016} and is the most sensitive vital sign marker that indicates clinical deterioration \cite{Schein1990, Goldhill1999, Ridley2005, Cretikos2008}. RR is therefore a highly informative indicator for the patient's health states and continuous monitoring of RR is desired.\\
 On the other hand, it is known that physiological mechanisms cause ECG and PPG signals to be modulated by respiration. Three types of modulation can be observed, as seen in Figure \ref{modulation}: 
 
\begin{figure}[h]
    \centering
\includegraphics[width=12cm]{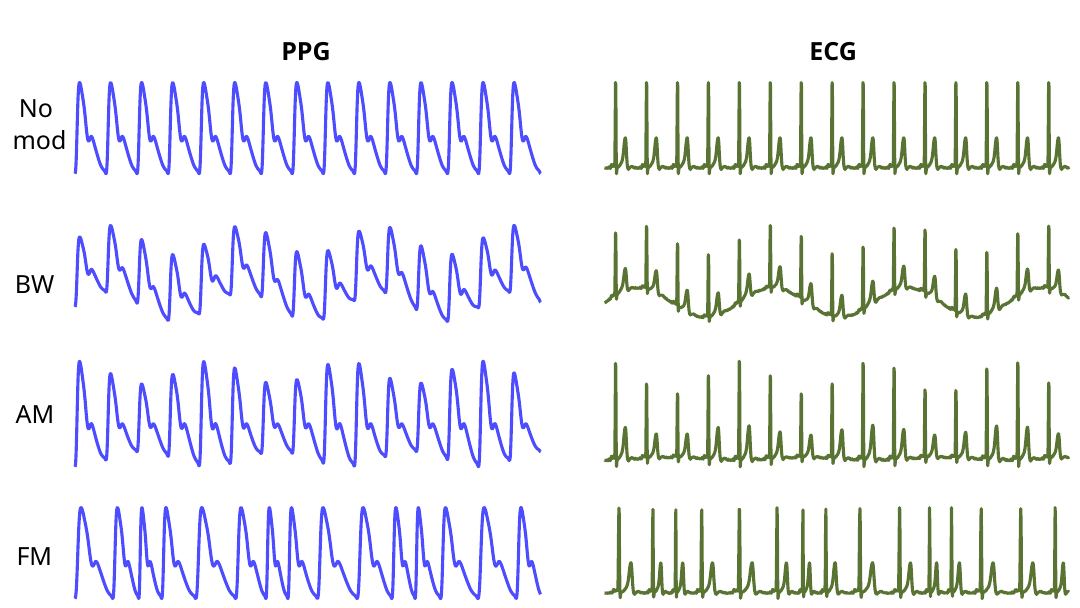} 
\caption{Idealised respiratory modulations of the PPG (left) and ECG (right). From top: no modulation, baseline wander (BW), amplitude modulation (AM), and frequency modulation (FM). Adapted from \cite{Addison2012} and \cite{Pimentel2015} by Peter H. Charlton - Own work, CC BY 3.0, \url{https://commons.wikimedia.org/w/index.php?curid=104501363}}
\label{modulation}
\end{figure}

Baseline wander (BW), amplitude modulation (AM) and frequency modulation (FM) \cite{Bailon2006, Meredith2012}.\\

 This motivates to try to gain insight on the respiratory rate from PPG-signals. Recently, approaches to estimate RR have been incorporated into consumer wearables which measure the PPG, and therefore it is important to investigate the performance of this approach, and to optimise it.

 \subsection{Potential datasets}
Besides the three datasets which are considered in Qumphy and are described below, i.e. MIMIC-III-Ext-PPG, MIMIC Perform Large and OSASUD, there are many other datasets containing PPG signals and respiratory signals or values for respiratory rate. Examples are the Vortal dataset \cite{Charlton2016}, BIDMC \cite{Pimentel2017} that is hosted on PhysioNet \cite{Goldberger2000}, and \href{https://borealisdata.ca/dataverse/capnobase}{Capnobase}. 

 \subsection{Data selection}
 MIMIC-III-Ext-PPG, MIMIC Perform Large and OSASUD are our preferred datasets for RR estimation based on the same assessment dataset rules as for the other benchmark problems.
 
\subsubsection{MIMIC-III-Ext-PPG}
The MIMIC-III-Ext-PPG dataset \cite{PhysioNet-mimic-iii-ext-ppg-1.1.0} was introduced in Sec.~\ref{sec:datasets_af}. For the subset of samples were a respiration (RESP) channel was present, we extracted also respiratory rates leveraging best practices from the literature \cite{charlton2021impedance}. This resulted in 4.3M 30s PPG waveform segments with respiratory rate annotation from more than 4600 patients.

 \subsubsection{MIMIC Perform Large}
MIMIC Perform Large is a newly created dataset which is an extension of the MIMIC Perform datasets that are extracted from the \href{https://physionet.org/content/mimic3wdb/1.0/}{MIMIC III Waveform Database}. On \href{https://zenodo.org/records/15906524}{zenodo}, a training and a test set are provided:\\

\begin{itemize}
    \item MIMIC PERform Large Training Dataset: 5,248 Recordings from 681 patients during routine clinical care, all of whom are adults. %
\item MIMIC PERform Large Testing Dataset: 1,257 Recordings from 167 patients during routine clinical care, all of whom are adults. %
\end{itemize}

In both datasets, recordings are 32 seconds long, with reference respiratory rates (denoted rr) derived from the imp signals using the algorithm proposed in \cite{charlton2021impedance}.

 \subsubsection{OSASUD}
 As pointed out before in the Sleep apnea benchmark problem, the OSASUD consists of 30 subjects, so each fold consists of nocturnal raw PPG signals and estimated features from 3 subjects. Segment duration - 1 min (raw PPG segment - 4800 samples, PPG feature sequence - 60 samples)
 
 \subsection{Making the datasets available}
MIMIC-III-Ext-PPG will be released as a dataset on Physionet along with a corresponding dataset descriptor to be submitted to Scientific data.\\
MIMIC Perform Large can be downloaded via \href{https://zenodo.org/records/15906524}{zenodo} where a training and a test set are provided. Code to create 10 stratified folds with respect to respiration rate and the 10 folds can be downloaded from our \href{https://gitlab.com/qumphy/d4-code/}{repository} very soon.\\
The original OSASUD dataset can be downloaded from \href{https://springernature.figshare.com/collections/A_dataset_of_stroke_unit_recordings_for_the_detection_of_Obstructive_Sleep_Apnea_Syndrome/5630890}{here}. Code to create 10 stratified folds with respect to respiration rate and the 10 folds can be downloaded from our \href{https://gitlab.com/qumphy/d4-code/}{repository} very soon.

\subsection{Data usage}
All three datasets, MIMIC-III-Ext-PPG, MIMIC PERform Large and OSASUD are provided together with splits into multiple folds. Thus there are many different possibilities for the use of the datasets: For every dataset, training, validation, calibration and test sets can be taken from the provided folds such that models could be calculated from MIMIC-III-Ext-PPG or MIMIC PERform Large or OSASUD respectively. The other two datasets could be used as external test sets.\\
Alternatively, the original training and test set for MIMIC PERform Large could be used.

\section{Conclusion}%
\label{sec:conclusion}

Photoplethysmographic signals are inexpensive and easy to collect and contain valuable information on the cardiovascular, respiratory, and autonomic nervous systems which is not yet routinely exploited. The report helps to enable the scientific and medical engineering community to create methods such that this information can be deduced from the signals. This is done by providing six exemplary benchmark problems of high clinical interest together with suitable benchmark datasets that are freely available or on demand, and explanations how to use these datasets to solve the Benchmark problems. We hope to support therefore the scientific progress in the field of PPG analysis by our report.

\begin{appendices}
\end{appendices}

\printbibliography

\end{document}